**Article title**

AIWR: Aerial Image Water Resource Dataset for Segmentation Analysis

**Authors**

Sangdaow Noppitak[a], Emmanuel Okafor[b], Olarik Surinta[c,*]

**Affiliations**

[a] Faculty of Science, Buriram Rajabhat University, 31000 Buri Ram, Thailand

[b] SDAIA-KFUPM Joint Research Center for Artificial Intelligence, King Fahd University of Petroleum and Minerals, Dharan 31261, Saudi Arabia

[c] Multi-agent Intelligent Simulation Laboratory (MISL) Research Unit, Department of Information Technology, Faculty of Informatics, Mahasarakham University, Mahasarakham, 44150 Thailand

**Corresponding author's email address and Twitter handle**

Email address: olarik.s@msu.ac.th (O. Surinta), Twitter: mrolarik

**Keywords**

Instance segmentation; Deep learning; Computer vision; Aerial image; Water management; Natural water body; Artificial water body

**Abstract**

Effective water resource management is crucial in agricultural regions like northeastern Thailand, where limited water retention in sandy soils poses significant challenges. In response to this issue, the Aerial Image Water Resource (AIWR) dataset was developed, comprising 800 aerial images focused on natural and artificial water bodies in this region. The dataset was created using Bing Maps and follows the standards of the Fundamental Geographic Data Set (FGDS). It includes ground truth annotations validated by experts in remote sensing, making it an invaluable resource for researchers in geoinformatics, computer vision, and artificial intelligence. The AIWR dataset presents considerable challenges, such as segmentation due to variations in the size, color, shape, and similarity of water bodies, which often resemble other land use categories. The objective of the proposed dataset is to explore advanced AI-driven methods for water body segmentation, addressing the unique challenges posed by the dataset's complexity and limited size. This dataset and related research contribute to the development of novel algorithms for water management, supporting sustainable agricultural practices in regions facing similar challenges.

## SPECIFICATIONS TABLE

| Subject | Computer Science |
|---|---|
| **Specific subject area** | Water body segmentation plays a crucial role in identifying water sources and managing the distribution of water resources to optimize their use in the agricultural sector |
| **Type of data** | Image (JPG format) and Raw (JSON format) |
| **Data collection** | Aerial water resource images were collected from the Bing Maps website in satellite format, representing high-resolution aerial data. A zoom level with a resolution of 1:50 meters was selected. The study area focuses on northeast Thailand, spanning 168,825.34 square kilometers, specifically targeting plains and agricultural regions. The dataset is designed to capture both natural and artificial water bodies, with each aerial image containing one or more water bodies. The Aerial Image Water Resource (AIWR) dataset includes aerial images along with corresponding ground truth files in JPG and JSON formats. |
| **Data source location** | Country: Thailand<br>Location: northeastern region<br>Latitude: +14° 14' to +18° 27'<br>Longitude: +101° 15' to +105° 35' |
| **Data accessibility** | Repository name: Mendeley Data<br>Data identification number: 10.17632/d73mpc529b.3<br>Direct URL to data: https://data.mendeley.com/datasets/d73mpc529b/3 |
| **Related research article** | S. Noppitak, S. Gonwirat, O. Surinta, Instance segmentation of water body from aerial image using mask region-based convolutional neural network, in: The 3rd International Conference on Information Science and Systems (ICISS), 2020: pp. 61–66. https://doi.org/10.1145/3388176.3388184 |

## VALUE OF THE DATA

- The Aerial Image Water Resource (AIWR) dataset presents unique challenges due to its limited sample size, comprising 800 images depicting both natural and artificial water bodies. Segmenting one or more water bodies within each aerial image is complicated by variations in color, shape, and size, which require expert knowledge in remote sensing for accurate identification. For non-experts, correctly delineating these features poses significant difficulties. To mitigate this challenge, the dataset includes ground truth annotations for each image, enabling researchers to conduct detailed visualizations and analyzes.
- The AIWR dataset is an interdisciplinary resource with applications across multiple fields, including geoinformatics, geographic information systems, computer vision, and artificial intelligence. Its broad scope enhances its relevance to a diverse range of research endeavors.

The dataset facilitates the analysis of satellite data, particularly in the RGB channel, the segmentation of both natural and artificial water bodies, and the exploration of the geographical landscape of Northeast Thailand. Additionally, it offers opportunities for developing and training novel algorithms aimed at improving the precision of instance segmentation in aerial images. The versatility of the dataset accommodates diverse analytical approaches, further highlighting its significance in cross-disciplinary research.

- The AIWR dataset is openly available, promoting collaboration among researchers and developers. It provides a crucial resource for advancing the application of artificial intelligence (AI) tools in analyzing water bodies, particularly within the context of aerial imagery.

## BACKGROUND

In the northeastern region of Thailand, a significant portion of the population depends on agriculture as their primary livelihood. Consequently, much of the area is dedicated to agricultural activities. However, the soil in this region is predominantly sandy, exhibiting poor water retention capabilities [1]. This condition has created an urgent need for the government to monitor water sources to ensure sufficient water supply for farmers and to support agricultural productivity [2]. This situation highlights the growing significance of water management research. In response, our research team has concentrated on segmenting water bodies, specifically utilizing aerial images captured in RGB channels. The primary data collection area covers the northeastern region of Thailand, characterized by extensive plains and agricultural land.

To construct the AIWR dataset, we intentionally collected a limited sample of aerial images to test the ability to achieve high accuracy with restricted data. The images feature diverse shapes, colors, and sizes of water bodies. Experts carefully reviewed and validated the ground truth annotations for precision. The ongoing challenge is to refine existing segmentation algorithms or develop new architectures that effectively address water body segmentation within the constraints of a small aerial image dataset.

## DATA DESCRIPTION

In 2019, we collected the Aerial Image Water Resource (AIWR) dataset using the Bing Maps website, focusing on the northeastern region of Thailand. The study area spans 168,825.34 square kilometers, located between latitudes: +14° 14' to +18° 27' and longitudes: +101° 15' to +105° 35', as illustrated in Fig. 1. The land in this region is predominantly used for cultivating key economic crops such as rice, sugarcane, cassava, and animal-feed corn, as well as for growing economic trees like rubber and eucalyptus [3,4]. Additionally, agriculture in the northeast relies heavily on rainwater and water sourced from the Mekong, Chi, and Mun rivers [5].

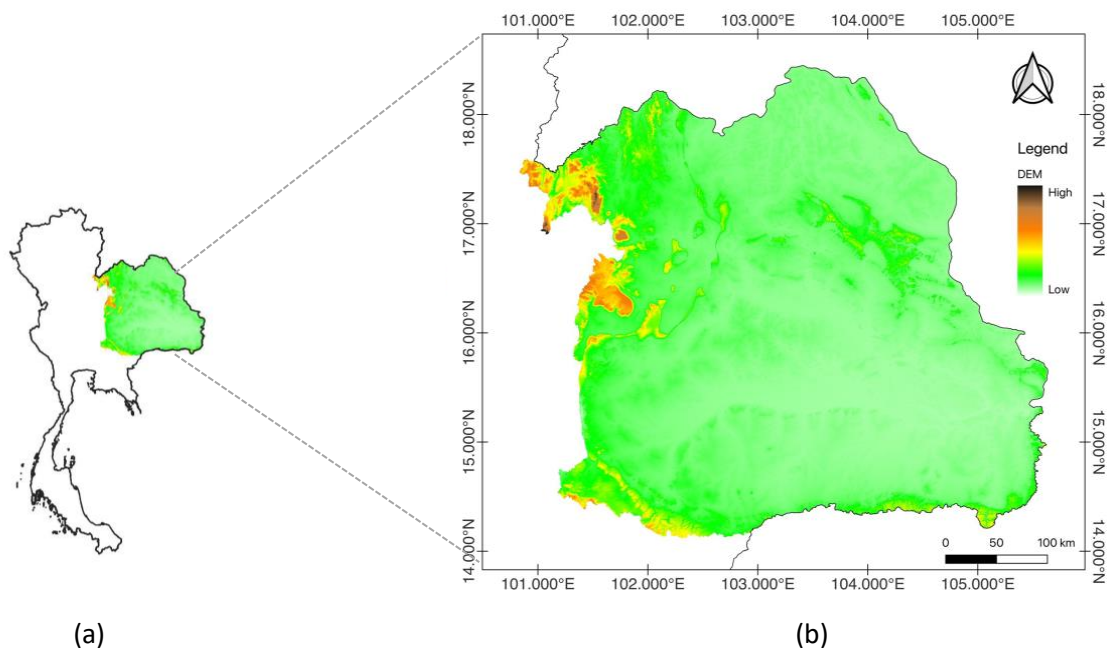

(a)                                        (b)

**Fig. 1.** Illustration of the study area in the northeastern region of Thailand: (a) the Thailand map and (b) the northeastern region.

The development of this dataset adheres to the standards set out in the Fundamental Geographic Data Set (FGDS). The land use layer is a critical component of Thailand's FGDS, developed as part of a project aimed at defining the content structure, attributes, and quality standards for the FGDS at the national level [6]. This initiative was led by the Geo-Informatics and Space Technology Development Agency (GISTDA) in collaboration with the Chulalongkorn University Academic Service Center. The content structure of the standard encompasses specific characteristics of FGDS layers, aligned with the guidelines outlined in the international standard ISO19131: Geographic Information – Data Product Specifications [6].

The land use classification system developed by the Land Development Department is the standard classification system for use across all government agencies [7]. In the AIWR dataset, the labelling of water body areas follows the Level 1 classification for land use in water body zones, categorized as "Water body (W)." The classification of water body land use is illustrated in Fig. 2.

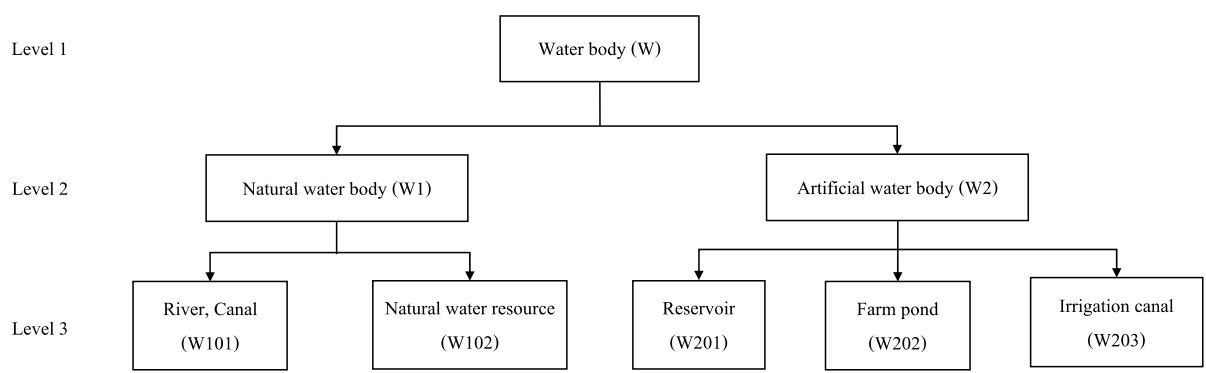

**Fig. 2.** The land use classification of water bodies is divided into three levels.

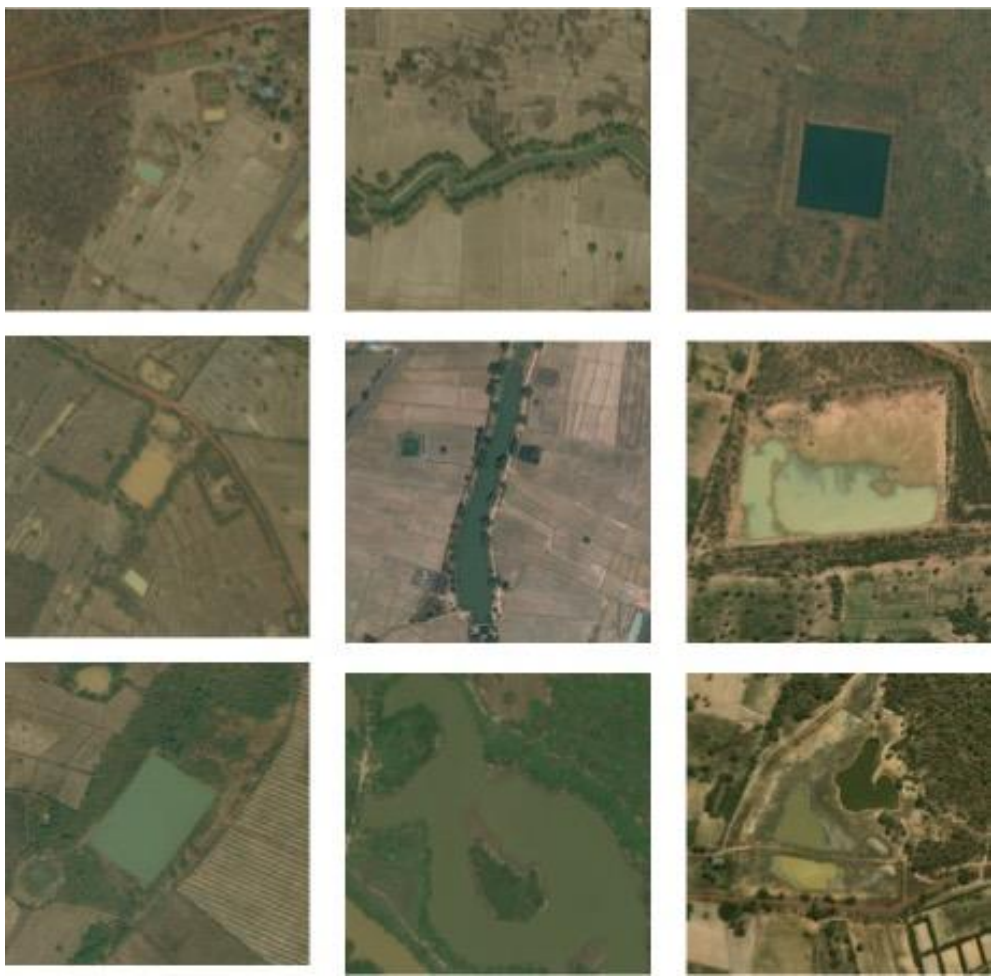

**Fig. 3.** Examples of water bodies (W), including both natural and artificial types.

The AIWR dataset consists of 800 aerial images that feature a diverse range of shapes, colors, and sizes of water bodies. Segmenting these water bodies is challenging, particularly for those who are new or intermediate in this field, as it requires significant time to study and accurately analyze the water body areas. The dataset includes both natural and artificial water bodies. Segmenting natural water bodies is particularly difficult due to the curved and irregular forms of rivers. Additionally, some areas exhibit varying colors, such as dark green, light green, brown, and dark blue (see Fig. 3).

The AIWR dataset was specifically curated to support researchers in the computer science domain, particularly those focused on developing advanced AI algorithms for water body segmentation in the northeastern region of Thailand.

The dataset presents significant challenges for instance segmentation due to variations in the size, color, and shape of water bodies. Additionally, the colors of these water bodies often closely resemble those of other land use categories, such as forests, flooded open areas, and paddy fields. The segmentation difficulties can be classified into four key aspects: color, shape, size, and similarity, as described below:

- **Color:** As shown in Fig. 4(a), the colors of water bodies exhibit considerable diversity, including shades of white, blue, gray, and black. In some locations, aquatic vegetation covers the water surface, resulting in darker hues like deep green and black.
- **Shape:** The shapes of water bodies vary widely, ranging from triangles and rectangles to U-shaped curves and zigzag patterns, as depicted in Fig. 4(b).
- **Size:** Water bodies also differ in size. Measurements from Bing Maps indicate that their widths range from 10 meters to 120 meters, as illustrated in Fig. 4(c). Notably, water bodies with a width of only 10 meters appear as mere small dots.
- **Similarity:** Aerial images of certain water bodies bear a striking resemblance to other land use types, such as flooded areas, water obscured by trees, or regions with built structures. Fig. 4(d) highlights these challenging areas using dashed lines to indicate regions with these similarities.

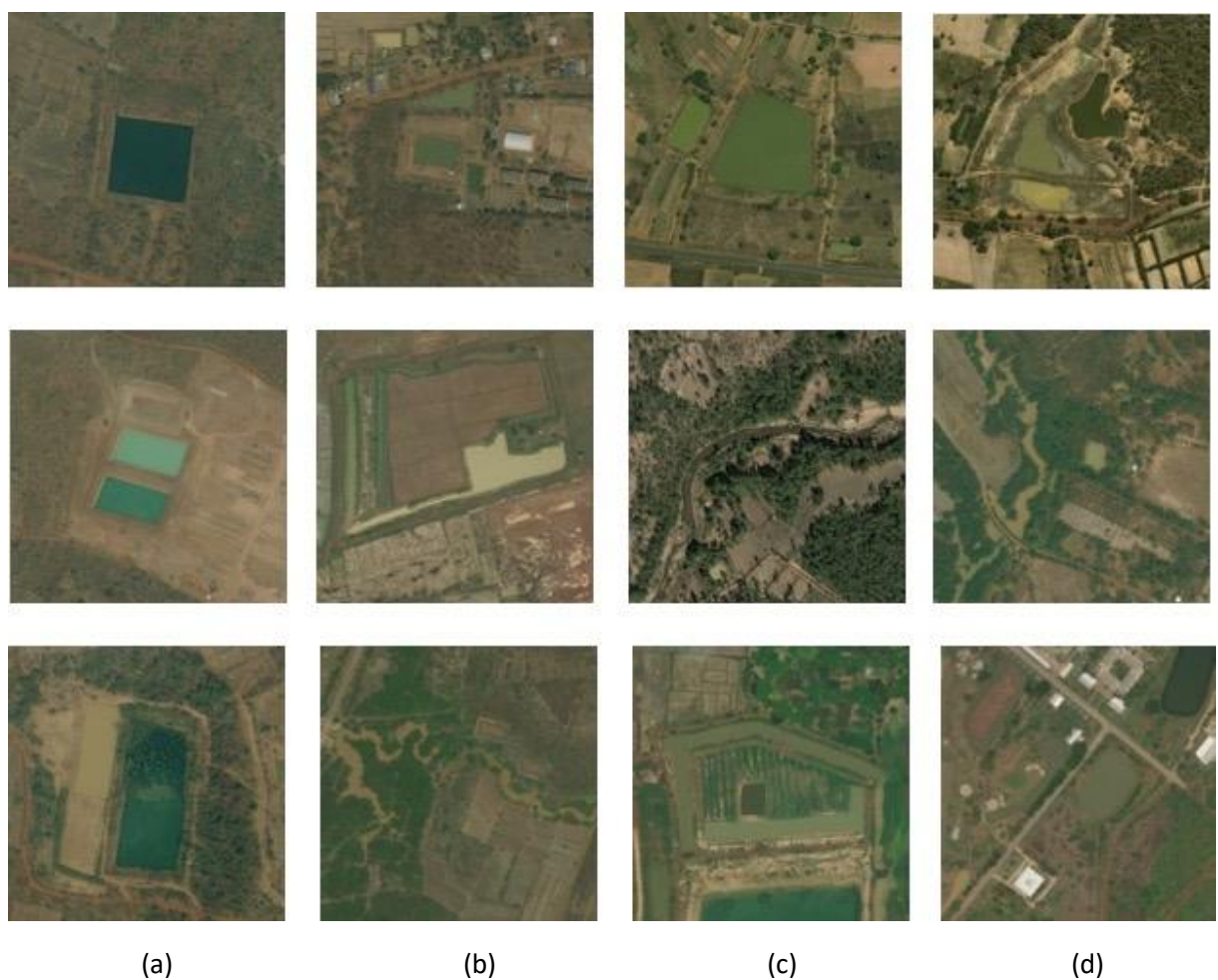

(a)  (b)  (c)  (d)

**Fig. 4.** illustrates the challenges of instance segmentation in the collected data, including (a) color, (b) shape, (c) size, and (d) similarity.

# EXPERIMENTAL DESIGN, MATERIALS AND METHODS

**Materials**

The aerial image data collected focuses on the northeastern region of Thailand and includes a limited dataset. This poses a challenge for researchers attempting to develop novel algorithms for segmenting water bodies in aerial images with a restricted training set.

The AIWR dataset is available and can be accessed through the Mendeley Data repository (https://data.mendeley.com/datasets/d73mpc529b/3) [8].

The dataset was created using Bing Maps in satellite format by randomly selecting land use areas classified as Level 1 (water body: W) in the northeastern region of Thailand. Each aerial image was designated to represent a single type of land use, specifically water bodies, with each image potentially containing one or more water bodies. The zoom level was set to achieve a resolution of 1:50 meters, as illustrated in Fig. 5. All aerial images were captured at a size of 650×650 pixels, maintaining a 1:1 aspect ratio.

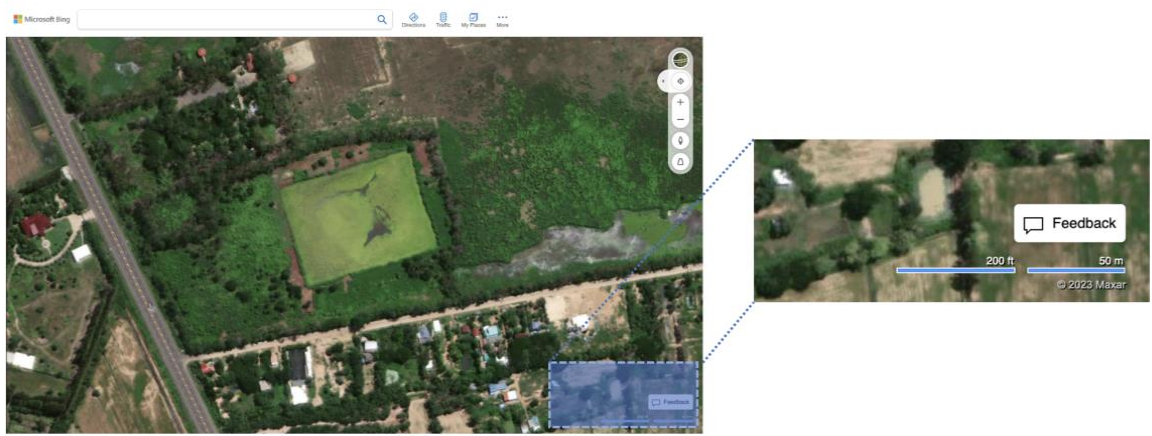

**Fig. 5.** Illustration of (a) Bing Maps at a zoom level with a resolution of 1:50 meters and (b) a closer zoom showing the specific level of detail.

Ground truth labels were assigned to each aerial image, with multiple labels possible within a single image, all categorized under the label "W." To ensure accurate delineation of water body boundaries, experts in remote sensing analyzed and interpreted the images. The ground truthing process was conducted using LabelMe software, which is available for download and installation from the following website: https://github.com/wkentaro/labelme/releases. Fig. 6 illustrates the screen capture of the LabelMe software and the labelling of the water body (W). In Fig. 6(a), the polygon points outlining the water body area are depicted. Fig. 6(b) provides a view of the labeled water body. Additionally, Fig. 6(c) displays the output generated by the LabelMe software, consisting of JPG and JSON files.

Fig. 7 presents examples of ground truth data that have been verified by experts.

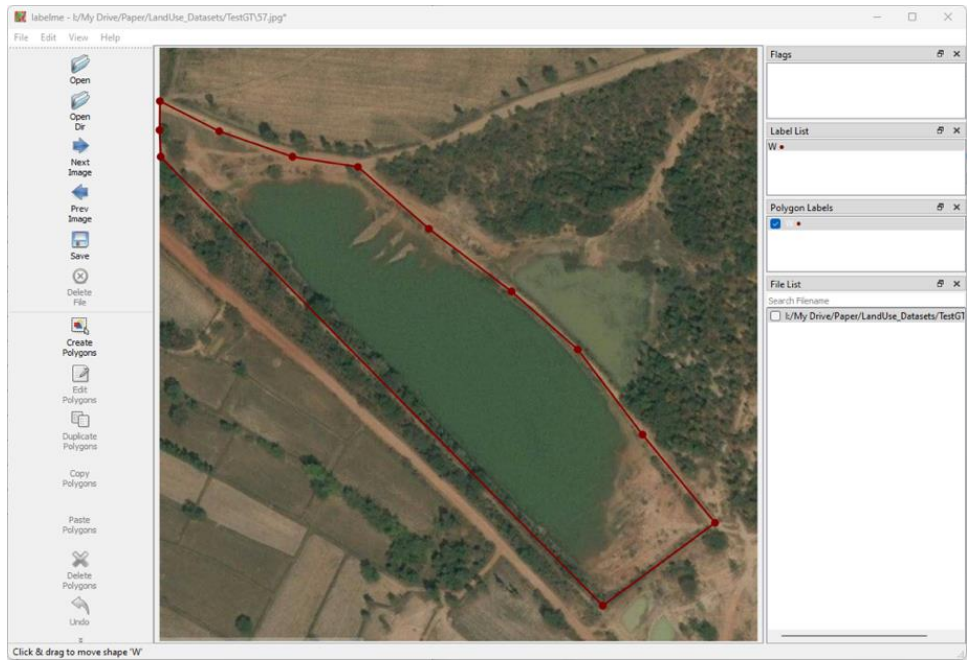

(a)

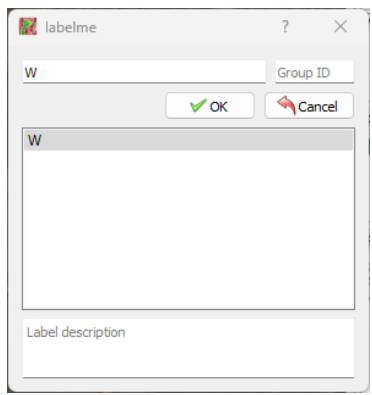

(b)

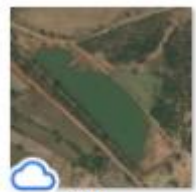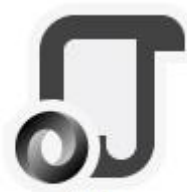

(c)

**Fig. 6.** Illustration of the LabelMe software: (a) main interface showing the generation of the water body area, (b) label assignment process, and (c) output files in JPG and JSON formats.

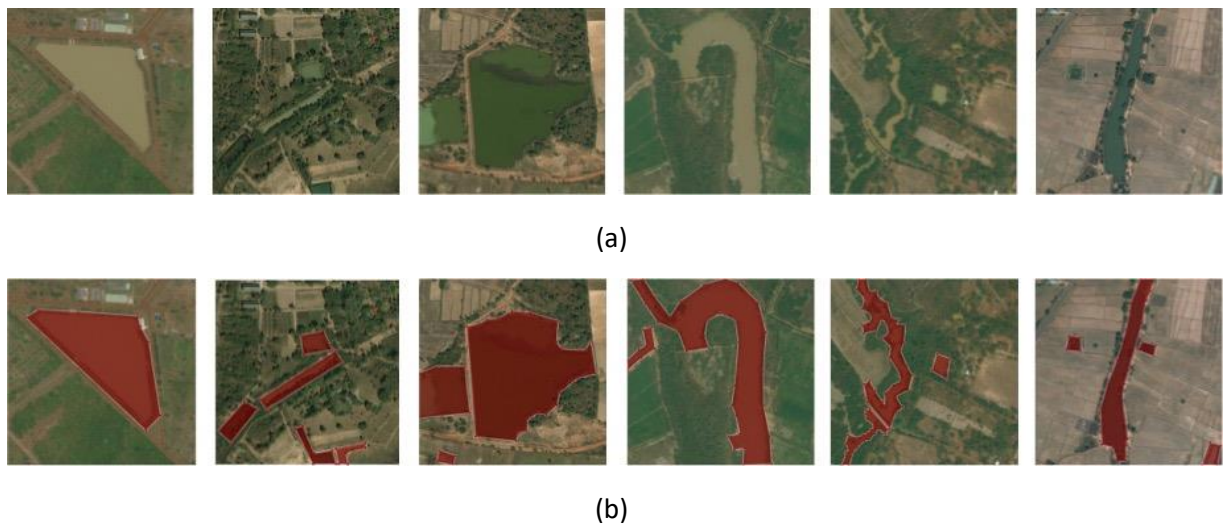

(a)

(b)

**Fig. 7.** Illustration of aerial images containing one or more water bodies: (a) the aerial images and (b) the corresponding ground truths.

The file structure of the dataset is illustrated in Fig. 8.

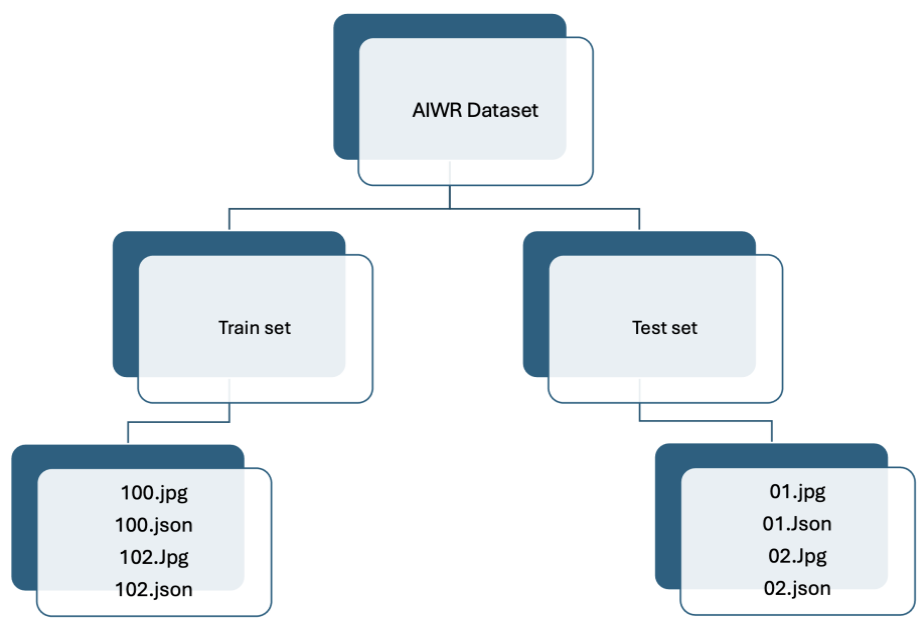

**Fig. 8.** Structure and format of files and filenames.

**Experimental design**

Using the AIWR dataset, Noppitak, Gonwirat, & Surinta [9] employed a mask region-based convolutional neural network (Mask R-CNN) with ResNet-101 as the backbone architecture. The dataset was divided into training, validation, and test sets in an 80:10:10 ratio. Data augmentation techniques were also applied during model training. The results indicated that the Mask R-CNN model achieved a validation loss of 0.41 and a mean Average Precision (mAP) value of 0.59.

In addition to selecting Mask R-CNN alongside other CNN architectures as the backbone, the experimental design also focuses on optimizing hyperparameters to enhance model performance. Hyperparameters such as learning rate, batch size, and the number of epochs are fine-tuned based on validation set performance [10]. Additionally, early stopping and learning rate scheduling are employed to prevent overfitting [11] and ensure that the model generalizes well across different test scenarios. The objective is to achieve a balanced trade-off between accuracy and computational efficiency, considering the limited dataset size and the complex nature of segmenting water bodies in the AIWR dataset.

Furthermore, several algorithms have been developed for segmenting areas of interest, including U-Net [12], PolarMask++ [13], YOLOv8 [14], and YOLACT++ [15]. These algorithms have demonstrated effectiveness across various research domains. However, the ongoing challenge lies in either refining these well-established segmentation techniques or developing a novel approach that can accurately segment water bodies despite the limitations posed by a restricted aerial image dataset.

## LIMITATIONS

Not applicable

## ETHICS STATEMENT

The authors have read and follow the ethical requirements and confirming that the current work does not involve human subjects, animal experiments, or any data collected from social media platforms.

## CRediT AUTHOR STATEMENT

**Sangdaow Noppitak:** Conceptualization, Data Curation, Investigation, Methodology, Resources, Validation, Writing – Original Draft; **Emmanuel Okafor:** Conceptualization, Validation, Writing – Review & Editing; **Olarik Surinta:** Supervision, Conceptualization, Experimental Design, Writing – Review & Editing, Funding Acquisition.

## ACKNOWLEDGEMENTS

This research project was financially supported by Mahasarakham University, Thailand.

## DECLARATION OF COMPETING INTERESTS

The authors declare that they have no known competing financial interests or personal relationships that could have appeared to influence the work reported in this paper.